# *Tahḏīb*: A Rhythm-Aware Phrase Insertion for Classical Arabic Poetry Composition

**Mohamad Elzohbi**
Department of Computer Science
University of Calgary
Calgary, Alberta, Canada T2N 1N4
`melzohbi@ucalgary.ca`

**Richard Zhao**
Department of Computer Science
University of Calgary
Calgary, Alberta, Canada T2N 1N4
`richard.zhao1@ucalgary.ca`

## Abstract

This paper presents a methodology for inserting phrases in Arabic poems to conform to a specific rhythm using ByT5, a byte-level multilingual transformer-based model. Our work discusses a rule-based grapheme-to-beat transformation tailored for extracting the rhythm from fully diacritized Arabic script. Our approach employs a conditional denoising objective to fine-tune ByT5, where the model reconstructs masked words to match a target rhythm. We adopt a curriculum learning strategy, pre-training on a general Arabic dataset before fine-tuning on poetic dataset, and explore cross-lingual transfer from English to Arabic. Experimental results demonstrate that our models achieve high rhythmic alignment while maintaining semantic coherence. The proposed model has the potential to be used in co-creative applications in the process of composing classical Arabic poems.

## 1 Introduction

In classical Arabic literature, poetry plays a central role since the pre-Islamic era, serving as a medium for storytelling, emotional expression, social and religious commentary, and language preservation. A defining characteristic of classical Arabic poetry is its strict adherence to metrical rules summarized in the theory of *ʿArūḍ* (Frolov, 2000). These rules dictate the rhythmic patterns that define each poetic meter, and any deviation from the standard meters or their accepted variations is traditionally considered a flaw. Such a verse is described as "broken" (مكسور) for being rhythmically invalid.

In contrast to the syllable-based scansion, the rhythmic patterns in the theory of *ʿArūḍ* are determined by a mora-based approach based on the arrangement of consonants and vowels (Frolov, 2000), which can be represented in a binary format, let's say a: '1' for a vocalized letter (*Mutaḥarrik*), and a '0' for an unvocalized letter (*Sākin*). The sequence of '1's and '0's forms a rhythmic pattern that is essential to the identity of Arabic verse, and it is used to classify the verse into one of the sixteen canonical meters. Determining these patterns requires more than surface syllable count as it requires an understanding of the granular phonological structure of the verse.

Recent advances in natural language processing and generation (NLP/G) have led to increased interest in computational approaches to Arabic poetry (Alyafeai et al., 2023). However, generating metrically valid verse that also preserves semantic coherence remains a significant challenge. A major barrier is the complexity of the Arabic script and the necessity of full diacritization to infer the rhythm accurately, a requirement unmet by most available corpora, which are only sparsely or inconsistently diacritized due to the natural tendencies of native Arabic speakers to omit "known" diacritics.

One of the main challenges, particularly for amateur poets, is expressing the intended meaning within the constraints of classical meters. The rhythmic structure restricts word choice and sentence construction, creating a tension between content and form that makes the writing process more difficult. Many modern poets opt for greater freedom in form, allowing meaning and emotion to guide their choices rather than strict metrical patterns in what is known in the Arabic literature as *al-Šiʿr al-Ḥurr* (free verse) (El-Azma, 1969; Al-Tami, 1993).

In this paper, we propose a rhythm-aware phrase insertion methodology for assisting in the composition of classical Arabic poetry. Our approach leverages ByT5 (Xue et al., 2022), a byte-level multilingual transformer model, which we fine-tune using a conditional denoising objective to enable it to insert or reconstruct phrases to align with a given rhythmic pattern. Our method is designed to function without requiring fully diacritized input during inference. Instead, the model learns to infer text that aligns with rhythmic patterns from zero to partially diacritized context. We adopt a curriculum learning strategy (Soviany et al., 2022) and explore cross-lingual transfer from a similar English lyrics generation task. We empirically demonstrate the benefits of curriculum learning in enhancing the model's ability to generate rhythmically valid verse. Our work has the potential to be used in co-creative tools that assist poets in composing classical Arabic poetry that adheres to specified rhythmic patterns, allowing authors to iteratively refine their poems with rhythmically valid suggestions, rather than generating entire verses automatically without human-in-the-loop supervision.

## 2 Related Work

Research on Arabic poetry processing has evolved over the past decades, moving from traditional rule-based approaches to machine learning and deep learning techniques. Early computational studies focused primarily on tasks such as meter classification and sentiment analysis, often relying on handcrafted linguistic rules and expert knowledge of classical Arabic prosody (Qarah, 2024).

With the advent of deep learning, particularly recurrent neural networks (RNNs) and transformer based architectures, there has been a notable shift toward data-driven approaches for Arabic poetry analysis and generation (Alyafeai et al., 2023). Recent works have leveraged pre-trained language models to generate Arabic poetry, aiming to improve fluency, coherence, and adherence to poetic conventions. For example, Beheitt and Hmida (2022) proposed an autoregressive approach in which GPT-2 (Radford et al., 2019) was first pre-trained on Arabic news from scratch, then fine-tuned on Arabic poetry. Abboushi and Azzeh (2023) adopted a similar approach where they started fine-tuning from the AraGPT2 (Antoun et al., 2021) parameters to complete Arabic poems showing promising results in fluency, coherence, meaning and meter and rhyme adherence. The *Ashaar* project (Alyafeai et al., 2023) provided a comprehensive framework for poetry analysis and conditional generation, including models for meter, era, and theme classification, as well as diacritization.

Despite these advances, most existing generation models either generate poetry from scratch or complete verses in an automated fashion without clear metrics to ensure the creativity of the generated text. In contrast, our work advocates for a co-creative approach to poetry generation, where human authors remain central to the creative process while receiving assistance in meeting the formal requirements of classical Arabic prosody. Moreover, while some models incorporate meter as conditioning signals, they are limited to a distribution based on the poetry corpus and the frequency of each meter as they do not integrate explicit transformations to ensure the relationship between the rhythm and the script is recognized.

Our work addresses these gaps by proposing a hybrid approach that combines the strength of transformer-based language models and rule-based methods. Specifically, we introduce a rhythm-aware phrase insertion framework by fine-tuning ByT5 using a conditional denoising objective. Our model leverages a rule-based grapheme-to-beat transformation to extract rhythmic patterns from the Arabic script, allowing a more explicit enforcement of desired rhythmic constraints specified by the users, even if they do not follow the most common meters or the traditional metrical patterns in general. Our methodology builds on our previous work on English lyrics generation (Elzohbi and Zhao, 2024), where we trained a ByT5 model to replace or insert words to align with a desired beat pattern. In this work, we extend this approach to classical Arabic poetry, addressing the unique orthographic and phonological features of Arabic script.

## 3 Methodology

We selected the ByT5 model, which builds upon the T5 (Text-to-Text Transfer Transformer) framework (Raffel et al., 2020). T5 is an encoder-decoder transformer designed for a variety of NLP tasks, with each task defined through a prompt prefix. Unlike the token-based models, ByT5 processes input at the character level, allowing for fine-grained control over character-level patterns.

### 3.1 Task Formalization

The task is formalized as inserting a set of words $W' = (w'_1, w'_2, \ldots, w'_i)$ into a poetry verse $S = (w_1, w_2, \ldots, w_n)$, such that $W'$ adheres to a given rhythmic pattern $G2B(W')$. We will refer to this task in the course of this paper as the *substitution task*. $G2B(.)$ is a Grapheme-to-Beat transformation function that converts a set of words into the rhythmic pattern as defined in the next section.

### 3.2 Grapheme-to-Beat Transformation

A fully diacritized Arabic script is typically moraic, implying a close correspondence between graphemes and their sounds. Nevertheless, there are exceptions that need to be processed (El-Imam, 2004). In Arabic prosody, the scansion process often rely on a systematic transcription called *al-Kitābah al-ʿArūḍīyyah* or *Taqtīʿ* (Frolov, 2000), which enforces a one-to-one mapping between diacritized graphemes and their corresponding consonant-vowel sequence and in turn the rhythmic pattern.

Assuming a fully diacritized Arabic script that includes *Hamzat al-Waṣl* (an often assimilated glottal stop) and marks silent graphemes, the grapheme-to-beat transformation can be performed using a rule-based method. These rules can be found scattered in traditional Arabic prosody books, such as in Al-Moqri and Al-Mubaraki (2009), and can be summarized by the following:

- **Process special known words:** This includes known words that are missing one of the long vowel graphemes, such as the singular feminine demonstrative pronoun (هَذِه), which is missing a long vowel grapheme, is replaced by (هَاذِهِي), fully diacritized with adding the missing long vowel grapheme. We compiled a dictionary of similar special words in our transformation.[1]

- **Expand the *Madda* letter:** which is a single grapheme (آ) that represents a glottal stop with a long vowel sound (/aː/). This must be expanded to (أَا) as separate graphemes.

- **Add *Išbāʿ*:** which is adding the missing long vowel grapheme that extends a vocalized letter at the end of a word. The addition can be either mandatory or optional, with the mandatory cases as follows:

    - A long vowel must be added to the pronoun clitics *hu* and *hi* when they are positioned between two vocalized letters. For example, *lahu mā* (لَهُ مَا) becomes *lahū mā* (لَهُو مَا) by appending the /uː/ sound to the pronoun.

    - A long vowel is required for the plural-*m* suffix when it is positioned between two vocalized letters and diacritized with a short vowel. For instance, *lahumu mā* (لَهُمُ مَا) becomes *lahumū mā* (لَهُمُو مَا) with the addition of the /uː/ sound.

    - A long vowel must also be added if a word appears at the end of a verse, has a vocalized ending, and is diacritized with a short vowel.

By default, the plural-*m* suffix is not vocalized. However, it is common practice to vocalize it when the rhythm require, this can be viewed as a poetic license in medial verse. In cases where the plural-*m* suffix is not marked with a short vowel diacritic, there is no certainty that the long vowel should be added. However, the addition of the long vowel follows the rhythm constrains only.

- **Expand Nunation (*Tanwīn*):** Replace (ً), (ٍ), and (ٌ) with (نَ), (نٍ), and (نٌ), respectively to include the final /n/ sound.

- **Expand Gemination (*Tašdīd*)** Replace the grapheme that has a gemination mark with two versions of the same grapheme, an unvocalized version followed by a vocalized version. For example, the verb (عَلَّمَ) meaning "he taught" becomes (عَلْلَمَ).

- **Remove Silent Graphemes:** Assuming that silent graphemes are marked with a special diacritic, these letters will be removed. For instance, the proper noun "*ʿAmr*" (عَمْرُو) becomes (عَمْرُ) by removing the silent (و) marked with the (ْ) diacritic.

- **Process *Hamzat al-Waṣl* (ٱ):**

    - *Case 1:* If it is found in the definite article (ٱل) followed by a sun letter (coronal consonant), remove the silent (ل) grapheme.

    - *Case 2:* If it appears at the beginning of a sentence, convert it to (أ) to indicate a glottal stop /ʔa/.

    - *Case 3:* If a vocalized letter precedes *Hamzat al-Waṣl*, remove *Hamzat al-Waṣl* as it will be silent in medial speech.

    - *Case 4:* If a long vowel precedes *Hamzat al-Waṣl*, remove both the vowel extension and the *Hamzat al-Waṣl*.

    - *Case 5:* If any unvocalized letter is followed by a *Hamzt al-Waṣl*, remove the *Hamzt al-Waṣl* and vocalize the unvocalized letter that preceded it.

---

[1] The source code, datasets and dictionaries used in this paper can be found here: https://github.com/melzohbi/poem-rhythm-arabic

After these transformations, each grapheme $g$ in an Arabic script sequence $S$ is paired with exactly one of four diacritic marks $d \in \{ \acute{\circ}, \acute{\circ}, \circ, \mathring{\circ} \}$. If $d \in \{ \acute{\circ}, \acute{\circ}, \circ \}$, we append '1' to the rhythmic sequence $G2B(S)$. If $d = \mathring{\circ}$, we append '0'.

### 3.3 Datasets and Preprocessing

To generate accurate rhythmic patterns by means of the rules described earlier from Arabic text, we require a fully diacritized script. However, most available Arabic texts are only partially diacritized or lack diacritics altogether. One possible approach would be to train a model to generate partially diacritized texts and then apply post-processing by means of a full-diacritization model for evaluation, but this introduces extra complexity. The available diacritization models are not perfect; even if they were, they lack some of the special diacritizations that are not commonly used such as *Hamzat al-Waṣl* and marking silent graphemes. Instead of the post-processing, we will train our model to generate fully diacritized outputs directly, but this will require a fully diacritized dataset for training.

We draw on the TASHKEELAH dataset (Zerrouki and Balla, 2017), which primarily contains Classical Arabic (CA) with some Modern Standard Arabic (MSA) examples. This dataset contains various text types from various books (e.g., religious, linguistic, literary, and news articles) annotated with various rate of diacritization. Because we aim to handle poetic text, we also utilize the APCD dataset (Yousef et al., 2019), which contains a substantial collection of Arabic poems across different eras, regions and types scraped from *al-Mawsūʿah al-Šiʿriyyah* (الموسوعة الشعرية), a poetry corpus compiled by the Department of Culture and Tourism in Abu Dhabi[2] and is available online through a search engine, and *al-Dīwān* (الديوان) which is an online corpus and a search engine for Arabic poetry.[3]

First, we processed the TASHKEELAH dataset by splitting the paragraphs into individual lines based on line boundaries. The APCD dataset was segmented into verses, with each verse consisting of two hemistichs combined into a single line. This resulted in $6{,}134{,}608$ lines from the TASHKEELAH dataset and $1{,}831{,}727$ verses from the APCD dataset. These samples exhibited varying lengths and varying degrees of diacritization. Next, we cleaned the text by removing any diacritics erroneously applied to non-Arabic letters and filtering out all non-Arabic characters (e.g., digits and symbols). We also discarded lines containing fewer than four words to ensure sufficient context.

| Diacritic | APCD | TASHKEELA |
|---|---|---|
| *fatḥah* | 463.9 K | 89.6 M |
| *ḍammah* | 142 K | 22.9 M |
| *kasrah* | 207.1 K | 38.2 M |
| *sukūn* | 141.5 K | 32.8 M |
| *tanwīn fatḥah* | 14.3 K | 1.7 M |
| *tanwīn ḍammah* | 14.4 K | 1.5 M |
| *tanwīn kasrah* | 20.5 K | 2 M |
| *tašdīd* | 64.5 K | 13 M |
| *Hamzat al-Waṣl* | 0 | 10 |
| *Ṣifr mustaṭīl* | 0 | 0 |
| Total Diacritics | 1 M | 202 M |
| Total Consonantals | 1.9 M | 297.3 M |

Table 1: Diacritics distribution in the APCD and TASHKEELA datasets.

Not all examples in the TASHKEELAH and APCD datasets were fully diacritized (see Table 1 for details) and some diacritizations were inconsistent. Inconsistencies include omission of default *Sukūn*, irregular diacritization, and the absence of diacritics for silent letters. To ensure compatibility with our grapheme-to-beat transformation, which requires fully diacritized text, we filter, clean, and normalize samples as follows:

- Find and diacritize well-known, unambiguous words.

- Only accept lines in which every word is diacritized, with at least 50% of the letters in each word are diacritized.

- Ensure a consistent order and place of diacritics and fix if the order is not correct. In cases of double diacritization, the gemination mark must precede any other diacritic. Any illegal double diacritization is removed. Also in case of *Tanwīn Fatḥa* it should precede the *Alif*, which means: any (اً) will be fixed to (اً).

#### 3.3.1 Spot-Checking:

Following the initial processing, we conducted a manual review by randomly selecting 250 examples from each of the processed dataset. This revealed that most missing diacritics were the default

---
[2]https://poetry.dctabudhabi.ae/#/poems
[3]https://www.aldiwan.net

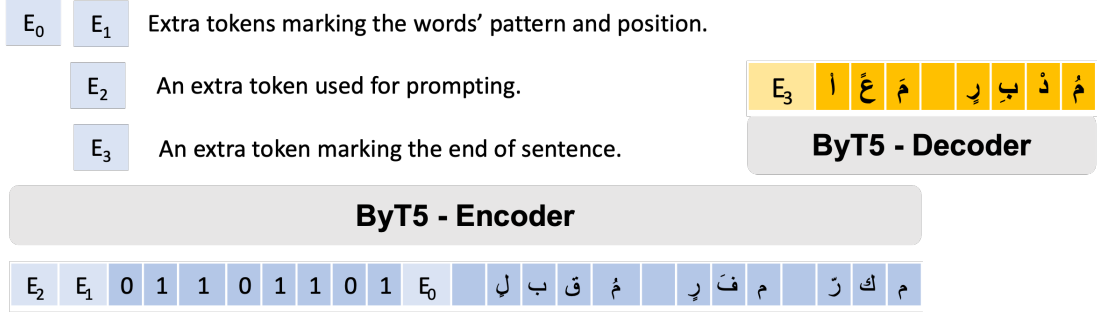

Figure 1: An example of the substitution task for Arabic text. The Arabic script, which is a single hemistich from a love poem composed by *Umru l-Qays* (†c. 544 CE), is displayed from right to left matching the order how it is display on the screen rather than how it is stored. It is displayed in non-cursive form for alignment purposes. The cursive form of the Arabic script in the input is: مكرٍّ مفرٍّ مُقبلٍ and in the output is: مُدْبِرٍ مَعًاْ.

*Sukūn* markers (indicating the absence of a vowel) and diacritics for silent letters and *Hamzat al-Waṣl*. To address the errors we noticed, we processed the dataset further as follows:

- An initial *Alif* if it appears at the beginning of a line, or follows a whitespace or a vocalized letter, and precedes an unvocalized letter or a gemination is most likely a *Hamzat al-Waṣl*. Similarly, the definite article (ال) under similar conditions. We change the non-diacritized *Alif* to *Hamzat al-Waṣl* in these cases.

- Adding a *Kasrah* diacritic to the (إ) letter, which is the only diacritic that can be applied to this letter.

- Marking silent letters with a special diacritic (these are silent *Alif*s in وا, used for the masculine plural at the end of a word, as well as in specific words such as مائة "meaning one hundred" and the proper noun عمرو. We will use the *al-Ṣifr al-Mustaṭīl* (ْ) diacritic to mark these silent letters.

- Assigning the default *Sukūn* diacritic to any remaining non-diacritized letters.

A second manual review was then performed on 250 randomly sampled examples from each dataset. In the APCD dataset, 204 lines were found to be error-free, 33 lines contained one error in one word, 11 lines contained errors in two words, and 2 lines contained errors in three words. Out of a total of 2,168 words, 63 words had errors, corresponding to a word error rate (WER) of $2.90\%$. Moreover, among $8,961$ diacritics, only 61 errors were observed, resulting in a diacritic error rate (DER) of $0.84\%$. Because our model samples from the data using a geometric distribution, the likelihood of selecting or retaining a word with an incorrect diacritic is very low. Even if some errors are picked up, the model is expected to learn to correct them probabilistically. Similar results were observed for the TASHKEELAH dataset.

Ultimately, we obtained $2,846,062$ fully diacritized lines from TASHKEELAH and $35,624$ from APCD. These datasets were then used to fine-tune our models for the *substitution task*, enabling them to learn the structures of diacritized Arabic in the context of poetic form and language.

### 3.4 Model Training

We fine-tuned a pretrained ByT5-base model on the task described earlier using the processed TASHKEELAH and APCD datasets. During training, we used a masking strategy to simulate the task's objective. Let $S = (l_1, l_2, \ldots, l_n)$ denote a fully diacritized sequence of Arabic script, where each $l_i$ consists of a grapheme accompanied by up to two diacritics (two only in the case of gemination). We randomly select a subset of words $W \subset S$ to be fully masked and used as prediction targets, where the length of $W$ is sampled from a geometric distribution with probability parameter $p = 0.2$. This allows the model to handle word segments of varying sizes, following a span-masking approach similar to Span-BERT (Joshi et al., 2020).

While the words in the masked sequence $W$ remain fully diacritized, the diacritics in the remainder of the sequence, $S \setminus W$, are reduced to mirror typical diacritization practices. Specifically, we remove all the special diacritics associated with silent letters as they are not commonly used. We then reduce the default *Sukūn* markers with a probability of 50% to reflect the tendency of Arabic speakers to not diacritize unvocalized consonants or long vowel extensions. For other common diacritics, we sample the number of diacritics to keep from a geometric distribution from 0 up to the total number of diacritics in the word with $p = 0.2$ to reflect the varying diacritization habits of Arabic speakers favoring little to no diacritization.

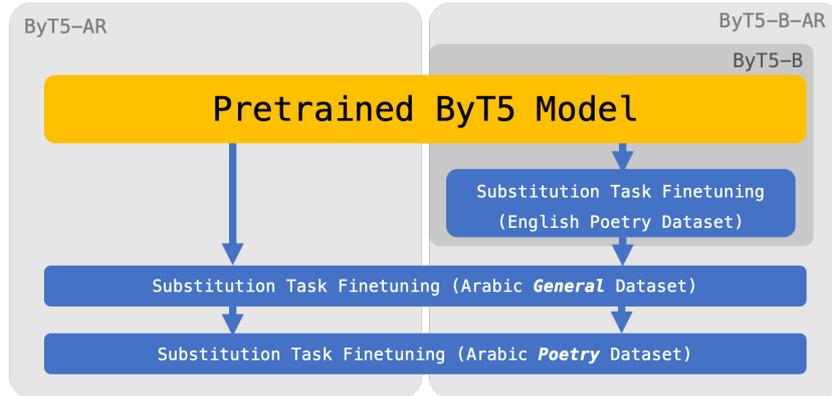

Figure 2: Illustration of the curriculum learning process for Arabic text.

Let $G2B(W)$ represent the rhythmic pattern corresponding to the masked target sequence of words $W$. We encapsulate $G2B(W)$ within special tokens $(E_0, E_1)$ and insert it in place of $W$ in $S$ to form a new sequence

$$S' = (l'_1, \ldots, E_0, G2B(W), E_1, \ldots, l'_n),$$

where each $l'_i$ is the letter after diacritic processing. A special token $E_2$ is then appended to prompt the model to predict the original target words $W$, thereby learning to align them with their corresponding rhythmic patterns.

By exposing the model to partially diacritized inputs while requiring fully diacritized outputs, we enable it to generate fully diacritized text from simulated, real-world patterns. The fully diacritized output can then be converted into its corresponding rhythmic pattern using the grapheme-to-beat transformation rules. Model performance is then evaluated by measuring the accuracy of the generated rhythmic pattern $G2B(W)$.

## 4 Experimental Setup

### 4.1 Dataset Split

Starting from the processed TASHKEELAH and APCD datasets, we sample 3500 lines from each dataset for evaluation during the first and second training phases. We used the remaining lines from the TASHKEELAH dataset for training in the first phase and from the APCD dataset for training in the second phase.

### 4.2 Training Setup

We adopted a two-stage training strategy: first, pre-training on TASHKEELAH followed by fine-tuning on APCD. APCD is a smaller and more complex dataset than TASHKEELAH as it contains poetic language. This progression in data complexity functions as a form of curriculum learning, since the poetic language in APCD presents a greater challenge than the more general and diverse language of TASHKEELAH.

In addition, we explored the potential benefits of cross-lingual knowledge transfer. To this end, we developed two models. The first model (referred to as **ByT5-B-AR**), is initialized with the parameters of the English lyrics generation model that we proposed in our previous work (Elzohbi and Zhao, 2024). This model was trained on a similar substitution task to generate English lyrics (referred to as **ByT5-B**), and then further fine-tuned on the Arabic substitution task using both TASHKEELAH and APCD. The second model, **ByT5-AR**, is initialized from the original **ByT5-base** and trained solely on the Arabic substitution task. Figure 2 illustrates the curriculum learning process employed in our experiments.

For both models, training was conducted for three epochs on the TASHKEELAH dataset, using a batch size of 128 for training and 16 for evaluation. Afterward, training continued for an additional three epochs on the APCD dataset with a reduced training batch size of 32 and evaluation batch size of 4. All experiments were executed on an NVIDIA A100 GPU with a learning rate of $3e - 4$ using a cosine scheduler and a weight decay of 0.01.

| First Training Phase on TASHKEELAH (3 epochs) | | | | |
|---|---|---|---|---|
| Evaluation Dataset | Model | Accuracy | Levenshtein | Coherence |
| Tashkeelah | ByT5-base | 26.31 | 79.06 | 29.63 |
|  | ByT5-AR | 71.86 | 95.41 | 29.43 |
|  | ByT5-B-AR | 72.31 | 95.35 | 29.37 |
| APCD | ByT5-base | 15.17 | 73.91 | 21.00 |
|  | ByT5-AR | 78.37 | 96.70 | **20.46** |
|  | ByT5-B-AR | **78.94** | **96.84** | 20.57 |
| Second Training Phase on APCD (3 epochs) | | | | |
| Evaluation Dataset | Model | Accuracy | Levenshtein | Coherence |
| Tashkeelah | ByT5-base | 41.37 | 87.23 | 28.88 |
|  | ByT5-AR | 73.00 | 95.36 | 29.08 |
|  | ByT5-B-AR | 73.06 | 95.28 | 29.13 |
| APCD | ByT5-base | 49.65 | 89.69 | 20.03 |
|  | ByT5-AR | 80.43 | 97.23 | **19.99** |
|  | ByT5-B-AR | **81.14** | **97.29** | 20.18 |

Table 2: Performance comparison of ByT5 models on the Arabic substitution task. The top section shows the results for models trained on the TASHKEELAH dataset (3 epochs), while the bottom section shows the results for models trained on the APCD dataset (3 epochs).

### 4.2.1 Automated Evaluation Metrics

To assess model performance, we use automated metrics adapted for Arabic. To measure the semantic coherence, we use mT5 (Xue et al., 2021), a multilingual variant of T5 that supports Arabic. Using its original span-denoising pretraining setup, we insert a special token at the masked span and prompt the model to predict the missing tokens. We then compute the cross-entropy loss between the mT5 predictions and those generated by our model.

$$loss(x, y) = -\log(\frac{e^{x_y}}{\Sigma_{i=1}^{n} e^{x_i}})$$

where $x$ is the logit output of the mT5 model's prediction, $y$ is the index of our model's predicted token in the mT5 vocabulary, and $n$ is the total number of tokens in the vocabulary. The loss is calculated per batch of 16 and averaged across all batches. Lower cross-entropy loss indicates better coherence as viewed by the pre-trained mT5 model.[4] All diacritics are removed from both the input texts and the model predictions to ensure consistency.

We also used the exact rhythmic alignment accuracy and the less restrictive Levenshtein similarity between the target and the generated rhythm as described in our previous work (Elzohbi and Zhao, 2024).

### 4.3 Experimental Results

Table 2 summarizes the performance of our models on the Arabic substitution task, evaluated in terms of rhythmic alignment and coherence.

After three epochs on TASHKEELAH, both **ByT5-AR** and **ByT5-B-AR** obtain comparable rhythmic alignment scores on both the TASHKEELAH and APCD evaluation sets, with **ByT5-B-AR** achieving slightly higher scores 72.31% and 78.94% than **ByT5-AR** 71.86% and 78.37%. Both models significantly outperform the baseline **ByT5-base** with scores of 26.31% and 15.17% on the TASHKEELAH and APCD evaluation sets, respectively.

Figure 3 shows that **ByT5-B-AR** begins with a higher baseline than **ByT5-AR**. This indicates that transferring knowledge from the English substitution task via curriculum learning (as in **ByT5-B-AR**) can accelerate early convergence for Arabic. However, the final performance gains from this cross-lingual transfer remain relatively modest.

Subsequent training on the APCD dataset for an additional three epochs further improves rhythmic alignment of our models by approximately 1

---
[4] we use the base-size version available at https://huggingface.co/google/mt5-base

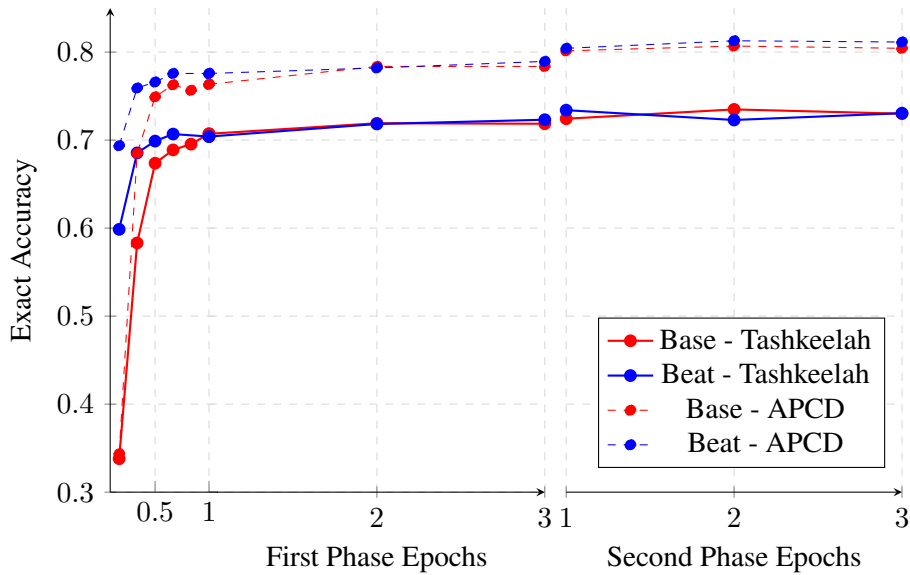

Figure 3: Exact accuracy of the ByT5 model on the Arabic substitution task.

point on the TASHKEELAH evaluation set and by 2 points on the APCD evaluation set. Interestingly, the training further enhanced the performance of the baseline **ByT5-base** model, which achieved a much higher improvements in accuracy especially on the APCD evaluation set with a $+34.48$ points improvement. We didn't notice signs of overfitting during the models' training, but it is possible that the base model learned how to adapt to the rhythmic pattern from the context without being explicitly exposed to the desired pattern as the APCD dataset is rhythmically structured. This demonstrates the advantages of training on structured poetic forms for the adaptation to the poetic domain. Nonetheless, these gains do not necessarily indicate a superior performance in generating poetic language. Human evaluation will be necessary to assess fluency and poetic qualities which we plan to conduct in future work.

All models exhibit similar coherence scores, suggesting that the fine-tuning process preserves semantic fluency while enhancing rhythmic alignment. Notably, the poetry-specific APCD evaluation set consistently achieves higher coherence and beat alignment scores compared to TASHKEELAH, even during the first training phase. This may be due to the consistent rhythmic structure of the APCD dataset and the use of full verses (two hemistichs) rather than individual lines, which likely provides a more sufficient context and thus supports improved coherence. Nonetheless, the high cross-entropy loss may also imply that the model lack decisiveness; an issue we aim to address through human evaluation.

## 5 Conclusion

In this paper, we investigated the capabilities of ByT5 for generating rhythm-constrained words in Arabic poems. Our methodology focused on fine-tuning ByT5-based models on a conditional denoising objective to reconstruct words with predetermined rhythmic patterns. Moreover, we validated our models using two diverse datasets: TASHKEELAH, which offers broad linguistic content, and APCD, characterized by a more structured poetic form. Our models showed high rhythmic alignment accuracy indicating their effectiveness in this task without adversely sacrificing the models' coherence based on automated evaluation metrics. Additionally, our experiments with cross-lingual transfer suggest that leveraging prior knowledge can accelerate early convergence, although the final performance gains are relatively modest, suggesting that the benefits of curriculum learning, especially in cross-lingual scenarios, may be inherently limited.

This model has a practical application in a co-creative rhythmic poetry composition framework. One limitation of our evaluation is that it relies on automated metrics, which may not fully capture the complex features of poetic language. To address this, we plan to conduct a human-centered evaluation to assess the fluency and poetic quality of the generated verses and its utility as a tool for assisting professional and amateur classical Arabic poetry composers.

# References

Omar Abboushi and Mohammad Azzeh. 2023. Toward fluent arabic poem generation based on fine-tuning aragpt2 transformer. *Arabian Journal for Science and Engineering*, 48(8):10537–10549.

Ismail Al-Moqri and Yahya A Al-Mubaraki. 2009. *Kitāb al-ʿArūd wa l-Qawāfī*. Transcription and Commentary, Dar Al-Nashr Lil-Jami'at.

Ahmed Al-Tami. 1993. Arabic" free verse": The problem of terminology. *Journal of Arabic Literature*, pages 185–198.

Zaid Alyafeai, Maged S Al-Shaibani, and Moataz Ahmed. 2023. Ashaar: automatic analysis and generation of arabic poetry using deep learning approaches. *arXiv preprint arXiv:2307.06218*.

Wissam Antoun, Fady Baly, and Hazem Hajj. 2021. Aragpt2: Pre-trained transformer for arabic language generation. In *Proceedings of the Sixth Arabic Natural Language Processing Workshop*, pages 196–207.

Mohamed El Ghaly Beheitt and Moez Ben Haj Hmida. 2022. Automatic arabic poem generation with gpt-2. In *ICAART (2)*, pages 366–374.

Nazeer Fowzi El-Azma. 1969. *Free verse in modern Arabic literature*. Indiana University.

Yousif A El-Imam. 2004. Phonetization of arabic: rules and algorithms. *Computer Speech & Language*, 18(4):339–373.

Mohamad Elzohbi and Richard Zhao. 2024. Let the poem hit the rhythm: Using a byte-based transformer for beat-aligned poetry generation. In *Proceedings of the 15th International Conference on Computational Creativity, (ICCC'24)*, pages 407–411.

Dimitry Frolov. 2000. *Classical Arabic Verse: History and Theory of ʿArūḍ*, volume 21. Brill.

Mandar Joshi, Danqi Chen, Yinhan Liu, Daniel S Weld, Luke Zettlemoyer, and Omer Levy. 2020. Spanbert: Improving pre-training by representing and predicting spans. *Transactions of the association for computational linguistics*, 8:64–77.

Faisal Qarah. 2024. Arapoembert: A pretrained language model for arabic poetry analysis. *arXiv preprint arXiv:2403.12392*.

Alec Radford, Jeffrey Wu, Rewon Child, David Luan, Dario Amodei, Ilya Sutskever, and 1 others. 2019. Language models are unsupervised multitask learners. *OpenAI blog*, 1(8):9.

Colin Raffel, Noam Shazeer, Adam Roberts, Katherine Lee, Sharan Narang, Michael Matena, Yanqi Zhou, Wei Li, and Peter J Liu. 2020. Exploring the limits of transfer learning with a unified text-to-text transformer. *Journal of machine learning research*, 21(140):1–67.

Petru Soviany, Radu Tudor Ionescu, Paolo Rota, and Nicu Sebe. 2022. Curriculum learning: A survey. *International Journal of Computer Vision*, 130(6):1526–1565.

Linting Xue, Aditya Barua, Noah Constant, Rami Al-Rfou, Sharan Narang, Mihir Kale, Adam Roberts, and Colin Raffel. 2022. Byt5: Towards a token-free future with pre-trained byte-to-byte models. *Transactions of the Association for Computational Linguistics*, 10:291–306.

Linting Xue, Noah Constant, Adam Roberts, Mihir Kale, Rami Al-Rfou, Aditya Siddhant, Aditya Barua, and Colin Raffel. 2021. mt5: A massively multilingual pre-trained text-to-text transformer. In *Proceedings of the 2021 Conference of the North American Chapter of the Association for Computational Linguistics: Human Language Technologies*, pages 483–498.

Waleed A. Yousef, Omar M. Ibrahime, Taha M. Madbouly, and Moustafa A. Mahmoud. 2019. Learning meters of arabic and english poems with recurrent neural networks: a step forward for language understanding and synthesis. *arXiv preprint arXiv:1905.05700*.

Taha Zerrouki and Amar Balla. 2017. Tashkeela: Novel corpus of arabic vocalized texts, data for auto-diacritization systems. *Data in brief*, 11:147.